\documentclass{article}
\usepackage{amssymb}
\usepackage{amsmath, amsfonts}
\usepackage{graphicx}
\usepackage{mathtools}
\usepackage[ruled]{algorithm2e}
\usepackage[T1]{fontenc}
\usepackage[dvipsnames]{xcolor}
\usepackage{subfig}
\usepackage{multirow}

\usepackage[preprint]{corl_2025} 

\newcommand{\reals}{\ensuremath{\mathbb{R}}}
\newcommand{\naturals}{\ensuremath{\mathbb{N}}}
\newcommand{\integers}{\ensuremath{\mathbb{Z}}}

\newcommand{\trans}{\ensuremath{\mathcal{T}}}
\newcommand{\expectation}{\ensuremath{\mathbb{E}}}

\newtheorem{theorem}{Theorem}

\definecolor{darkgreen}{rgb}{0.0, 0.5, 0.0}  

\title{Efficient Environment Design for Multi-Robot Navigation via Continuous Control}

\author{
  Jahid Chowdhury Choton\\
  Department of Computer Science\\
  Kansas State University, 
  United States\\
  \texttt{choton@ksu.edu} \\
  \And
  John Woods\\
  Department of Computer Science\\
  Kansas State University, 
  United States\\
  \texttt{jwoods03@ksu.edu} \\
  \AND
  William Hsu\\
  Department of Computer Science\\
  Kansas State University, 
  United States\\
  \texttt{bhsu@ksu.edu} \\
}

\begin{document}
\maketitle


\begin{abstract}
   Multi-robot navigation and path planning in continuous state and action spaces with uncertain environments remains an open challenge. Deep Reinforcement Learning (RL) is one of the most popular paradigms for solving this task, but its real-world application has been limited due to sample inefficiency and long training periods. Moreover, the existing works using RL for multi-robot navigation lack formal guarantees while designing the environment. In this paper, we introduce an efficient and highly customizable environment for continuous-control multi-robot navigation, where the robots must visit a set of regions of interest (ROIs) by following the shortest paths. The task is formally modeled as a Markov Decision Process (MDP). We describe the multi-robot navigation task as an optimization problem and relate it to finding an optimal policy for the MDP. We crafted several variations of the environment and measured the performance using both gradient and non-gradient based RL methods: A2C, PPO, TRPO, TQC, CrossQ and ARS. To show real-world applicability, we deployed our environment to a 3-D agricultural field with uncertainties using the CoppeliaSim robot simulator and measured the robustness by running inference on the learned models. We believe our work will guide the researchers on how to develop MDP-based environments that are applicable to real-world systems and solve them using the existing state-of-the-art RL methods with limited resources and within reasonable time periods.
\end{abstract}



\section{Introduction}
\label{sec:intro}

Developing robots that can perform challenging tasks in complex, uncertain environments is a fundamental problem in artificial intelligence. One of the most popular paradigms for solving such tasks has been the use of reinforcement learning (RL) algorithms based on the Markov Decision Process (MDP) formalism \cite{salimans2017, dulacArnold2021, sutton2018reinforcement}. But the progress in applying RL algorithms to real-world systems with continuous control has been very limited, and there is a lack of available benchmarks that represent multi-robot navigation via continuous control \cite{brockman2016openai, islam2017reproducibility, mania2018ars, dulacArnold2021}. Most RL researchers are focused on developing new RL methods, while the formal design and implementation of complex environments with continuous control are equally crucial and come with their own unique challenges. Developing robust, scalable, and effective RL environments requires deep understanding and careful design choices \cite{salimans2017, dulacArnold2021, sutton2018reinforcement}. Moreover, MDP-based environments can also be used to evaluate non-gradient or even non-RL based methods such as Augmented Random Search (ARS) \cite{mania2018ars}, Evolution Strategies (ES) \cite{salimans2017}, Black Box Optimization \cite{risi2015neuroevolution}, etc.

In this paper, we present a formal analysis of designing an efficient and customizable environment for multi-robot navigation via continuous control. The goal is to visit a set of regions of interest (ROIs) using a fixed number of robots. We formulate the multi-robot navigation task as an optimization problem, and formally model the environment as a Markov Decision Process (MDP). The state space is a continuous multi-dimensional space representing the robots' positions and dynamics, and the action space is a continuous 2-D space representing the forces applied to the robots to change their velocities. Each state has the positions and velocities of each robot and information about how many regions of interest (ROIs) are visited. The information about the visited ROIs is necessary to move towards the unvisited ROIs by taking optimal action. This makes the state space complete and observable compared to the Partially Observable Markov Decision Process (POMDP) \cite{meng2021memory}. But this also makes the state space much larger and more challenging to find optimal solutions. We crafted ten different variations of the environment and measured the performance of five state-of-the-art gradient-based deep RL algorithms: Advantage Actor-Critic (A2C) \cite{mnih2016a2c}, Trust Region Policy Optimization (TRPO) \cite{schulman2015trust}, Proximal Policy Optimization (PPO) \cite{schulman2017ppo}, Truncated Mixture of Continuous Distributional Quantile Critics (TQC) \cite{kuznetsov2020tqc}, CrossQ \cite{bhatt2024crossq} and a finite-difference based RL algorithm named Augmented Random Search (ARS) \cite{mania2018ars}. As \citet{islam2017reproducibility} and \citet{henderson2018deep} stated the importance of measuring the performance of different methods over random seeds, we used the hyperparameter optimization library Optuna \cite{optuna_2019} to select a set of hyperparameters for each algorithm. To show the real-world application of our environment, we deployed the environment on an agricultural field consisting of herbicide spraying robots using CoppeliaSim \cite{coppeliaSim2013}, where the goal is to selectively spray herbicides on the weed-infected locations. We included winds blowing in a certain direction in the environment to introduce uncertainty and measured the performance of the trained models to assess their robustness. The experiments show that our environment is learnable within a reasonable training period and samples, despite having a large, infinite, and continuous state and action space. We believe that our work can be used as a baseline to bridge the gap between developing efficient methods for multi-robot navigation and their application to real-world dynamic and complex environments.





\section{Related Work}
\label{sec:related}
Many efficient methods for multi-robot navigation and control have been developed in the last two decades \cite{nafis2021, sam2005, mickey2021, ratan2020, almadhounSurvey2019, lavalle2006planning, choton2023icra}. Classical methods such as A* \cite{goto2003}, RRT \cite{rodriguez2006rrt}, and RRT* \cite{karaman2011rrts} often rely on discrete planning techniques with a static environment and limited state space. Other techniques, such as Particle Swarm Optimization (PSO) \cite{nafis2021} and RA-MCPP \cite{mickey2021} also face difficulties in high-dimensional spaces due to their reliance on global optimization and often require explicit modeling of the path-planning heuristics. Therefore, multi-robot path planning in high-dimensional continuous state and action spaces remains a critical problem in robotics. RL has emerged as a promising approach due to its ability to learn optimal policies through interaction with environments \cite{sutton2018reinforcement}. While there has already been a few attempts at applying RL to multi-robot navigation \cite{kober2014, bae2019as}, these works have limited and discrete state and action spaces. 

\citet{dulacArnold2021} provided a series of independent real-world challenges that are holding back RL from real-world use, developed the corresponding MDPs, and measured their performances with two state-of-the-art RL algorithms. However, none of the challenges involve multi-robot navigation and control, and the formal analysis of the MDPs is limited and incomplete. Other interesting works on real-world environments include Google Research Football \cite{kurach2020google} (image-based) and CityFlow \cite{zhang2019cityflow} (non-image-based), but the formal analysis is again not present. To the best of our knowledge, there are no publicly available implementations or benchmarks that describe the MDP-based environment for multi-robot path planning via continuous control in a continuous state space, which can be tested with both RL and non-RL based methods for finding optimal navigation paths. \citet{mania2018ars} have stated that it is important to add more baselines that are extensible and reproducible because many RL algorithms that perform better in limited state and action space can perform significantly worse in high-dimensional continuous spaces \cite{dulacArnold2021}.



\section{Background}
\label{sec:background}
Let $\reals, \reals^+, \naturals, \integers$ be the set of reals, positive reals, naturals, and integers, respectively. For a set $S$, let $|S|$ denote its length. For two points $p = (p_1, p_2, ..., p_n) \in \reals^n $ and $q = (q_1, q_2, ..., q_n) \in \reals^n$ in $n$ dimensional space, the Euclidean distance \citep{euclid1956} is denoted as, $||p-q|| = \sqrt{(p_1-q_1)^2+(p_2-q_2)^2+...+(p_n-q_n)^2}$. Given two real numbers $x,y \in \reals$, let $[x,y] = \{ i \in \reals \mid x \leq i \leq y \}$ denote the set of all real numbers from x to y. Let $[n_{min}, n_{max}]^{p\times q}$ denote the set of matrices of shape $p \times q$ where every entry is clipped between $n_{min}$ and $n_{max}$ ($n_{min} \leq n_{max}$). Let $Y$ be a continuous random variable with probability density function $f$. The probability of getting the value $y$ for $Y$ is denoted as $P(Y=y)$. The expected value (or expectation) is defined as, $\expectation[Y] = \int_{-\infty}^{\infty} y f(y)dy$. For a set of samples $S$, let $\Delta(S)$ denote the set of probability distributions over $S$. Given a scalar-valued differentiable function of $n$ variables $f(x_1, x_2, ..., x_n)$, the gradient of $f$ with respect to its variables is defined as, $\nabla f = ( \frac{\partial f}{\partial x_1}, \frac{\partial f}{\partial x_2}, ..., \frac{\partial f}{\partial x_n})$. Using these notations, we will formally describe the multi-robot path planning problem in the next section.
\section{Problem Formulation}
\label{sec:problem}
A group of mobile robots operates in this continuous state space with position and velocity components along both the x and y axes. The control inputs correspond to forces applied along these axes. Let the environment for multi-robot navigation $F \subset \reals^2$ be a continuous 2-D space. For $n \in \naturals $ robots, let $(p_1,p_2,...,p_n) \in \reals^{2n}$ denote their positions in the environment where each $p_i=(x_i,y_i)$ denotes the coordinate of the center of the $i^{th}$ robot. Let $(v_1, v_2, ..., v_n)$ denote the corresponding velocities where each $v_i = (v_{x_i},v_{y_i})$ denotes the velocity of the $i^{th}$ robot along the x and y axis respectively. Let $W = (w_1,w_2,...,w_m) \in \reals^{2m}$ be a sequence of $m \in \naturals$ regions of interest where each $w_i = (w_{x_i}, w_{y_i}) \in F$ denotes the center of that region. Let $b_w(t) \in  \{0, 1, ..., (2^{|W|}-1)\}$ be the number that encodes how many regions are visited by the robots at time $t$. For example, if the first and third region is visited among five regions of interest, its binary representation is $10100$ which corresponds to a decimal value of $20$. If all five regions are visited, we have $11111$ which corresponds to a decimal value of $2^5-1=31$. Therefore, at time $t$ we have the state representation:
\[  s_t = ((p_1, p_2 ..., p_n), (v_1,v_2,...,v_n), b_w(t)) \]
Let $f_{x_{min}}, f_{y_{min}}, f_{x_{max}}$ and $f_{y_{max}}$ be the minimum and maximum forces that can be applied to change the velocity of a robot along the x and y axis respectively. Let $u_i(t) = (f_{x_i}, f_{y_i})$ be the control signal for the $i^{th}$ robot at time $t$. Therefore, the action consists of $n$ pairs of forces along the x and y axis:
\[ a_t = (u_1(t), u_2(t), ..., u_n(t)) = ((f_{x_1}, f_{y_1}), ...,(f_{x_n}, f_{y_n}))\]
Where each $f_{x_i} \in [f_{x_{min}}, f_{x_{max}}]$ and $f_{y_i} \in [f_{y_{min}}, f_{y_{max}}]$ is the force that is applied to change the velocity of the $i^{th}$ robot along the x and y axis, respectively. By applying $f_{x_i}$ and $f_{y_i}$ to the robot, the velocities are updated by:
\begin{flalign*}
    v'_{x_i} = v_{x_i} + \frac{f_{x_i}}{M_i} \\
    v'_{y_i} = v_{y_i} + \frac{f_{y_i}}{M_i} 
\end{flalign*}
Here, $M_i$ is the mass of the $i^{th}$ robot. If we have a windy environment where the wind is blowing at an angle $\beta_a$ and magnitude $v_a$, then we have the updated velocities:
\begin{flalign*}
    v'_{x_i} = v_{x_i} + \frac{f_{x_i}}{M_i} + v_a \cos\beta_a \\
    v'_{y_i} = v_{y_i} + \frac{f_{y_i}}{M_i} + v_a \sin \beta_a
\end{flalign*}
The positions are updated by:
\begin{flalign*}
    x_i' = x_i + v'_{x_i}\tau \\
    y_i' = y_i + v'_{y_i}\tau
\end{flalign*}
Where $\tau \in \reals^+$ is the time it takes to update the positions. Let $\trans: S \times A \to S$ be the transition function that determines the next state $s_{t'}$ by taking action $a_t$ from the state $s_t$ using the dynamics defined above where $|t'-t|=\tau$ and $s_{t'} = \trans(s_t,a_t)$. Let all robots be identical, and $C \in \reals^+$ be the minimum safety distance between any two robots to avoid collisions. Our goal is to visit the regions of interest $W$ using the $n$ robots by following the minimum path and avoiding collisions.
This can be achieved by solving the following optimization problem:
\begin{align}
& \min \sum_{i=1}^{n} \bigg\lvert \int_{t=0}^{T} u_i(t)dt \bigg\rvert   \label{eq:opt} \\
\textrm{subject to,    } & \nonumber \\
& b_w(T) = 2^{|W|} - 1 \nonumber \\
& \forall t \in [0,T], \forall i,j \in \{1,2,...,n\}: i\neq j \implies ||p_i-p_j|| > C \nonumber \\
& \forall t \in [0,T]: f_{x_i} \in [f_{x_{min}}, f_{x_{max}}], f_{y_i} \in [f_{y_{min}}, f_{y_{max}}], u_i(t)=(f_{x_i},f_{y_i}) \nonumber  \\
& \forall t \in [0,T]: s_{t+\tau} = \trans (s_t, a_t) \nonumber
\end{align}
where the transition function $\trans$ follows the robot dynamics described above, and $T$ is the total flight time. The first constraint ensures that all regions of interest (ROI) are visited. The second constraint ensures that no collisions have occurred. The third constraint ensures that at each time, the forces applied are within the maximum and minimum limits. The fourth constraint ensures that the robots are following the dynamics described above. This optimization problem can be represented by an MDP $M = (S, A, \trans, R, \gamma)$ where,
\begin{itemize}
\item 
$S \subset \reals^{4n+1}$ is the set of states where $s_t \in S$ at time $t$ is defined as above.

\item 
$A \subset \reals^{2m}$ is the set of actions where $a_t \in A$ at time $t$ is defined as above.

\item 
$\trans: S \times A \to S$ is the transition function that determines the next states as defined above.

\item 
$R: S \times A \to \reals$ is the reward function which is the sum of four different functions $R_C: S \times A \to \reals$, $R_I: S \times A \to \reals$, $R_F: S \times A \to \reals$ and $R_V: S \times A \to \reals$ that at time $t$ satisfies:
\[  R(s_t,a_t) = R_C(s_t,a_t) + R_I(s_t,a_t) + R_F(s_t,a_t) + R_V (s_t,a_t) \]
Let $r_s, r_m, r_l \in \reals^+$ be some arbitrarily fixed small, medium, and large rewards, respectively, where $r_s < r_m < r_l$. The first reward function $R_C$ determines the reward for any collisions between the robots:
\begin{align*}
&R_C(s_t, a_t) = \\
&   \begin{cases}
    -\infty &\text{if } \exists i,j \in \{1,2,...,n\}: (i\neq j) \land (||p_i-p_j|| \leq C) \\
    0, &\text{otherwise}
    \end{cases}
\end{align*}
The second reward function $R_I$ determines the reward that the robots get by visiting any region of interest (ROI). Therefore for some regions of interests $W_v \subseteq W$:
\begin{align*}
&R_I(s_t, a_t) = \\
&    \begin{cases}    \
    +\infty &\text{if } b_w(t) = 2^{|W|} - 1 \\
    \sum_{w \in W_v} r_l, &\text{if } \forall w\in W_v, \exists i\in \{1,2,...,n\}: p_i = w  \\
    0, &\text{otherwise}
    \end{cases}
\end{align*}

The third reward function $R_F$ determines the reward that the robots get by checking if their locations are within the field $F$ and get a large negative reward if not. Every robot will also get a small negative reward by moving into any location inside the field to minimize path length. Therefore for some set of robots $N_f$ where $|N_f|\leq n$:
\begin{align*}
& R_F(s_t, a_t) =\\
& \begin{cases}
    -\sum_{i \in N_f} r_l, &\text{if } \forall i \in N_f: p_i \notin F \\
    -r_s, &\text{otherwise}
    \end{cases}
\end{align*}

The fourth reward function $R_V$ determines the reward that the robots get by checking if the new state was visited previously in the current episode or not. Therefore:
\begin{align*}
& R_V(s_t, a_t) =\\
& \begin{cases}
    -r_m, &\text{if $s_{t'} = \trans(s_t,a_t)$ was visited before}\\
    r_s, &\text{otherwise}
    \end{cases}
\end{align*}

\item 
$\gamma \in [0,1)$ is the discount factor that is applied to future rewards.
\end{itemize}
A policy $\pi: S \to \Delta (A)$ is a function that maps a state to a probability distribution over the set of actions. The value of a policy $\pi$ at state $s$ is defined as the cumulative reward from following the policy starting from state $s$ and written as, $V^{\pi}(s) = \expectation_{\pi}[\sum_{t\geq0}\gamma^t r_t | s_0=s]$. The action-value or Q-value of a policy $\pi$ at state $s$ and action $a$ is defined as the expected cumulative reward from following the policy after taking the action $a$ starting from $s$ and written as, $Q^{\pi}(s,a) = \expectation_{\pi}[\sum_{t\geq0}\gamma^t r_t | s_0=s, a_0=a]$. In policy gradient methods \citep{sutton2018reinforcement}, a parameterized policy $\pi_{\theta}$ is determined by the parameter vector $\theta \in \reals^d$ which at time $t$ can be written as $\pi_{\theta}(a|s) = P(a_t=a | s_t=s, \theta_t = \theta)$. The goal is to maximize the expected return $J(\theta) = \expectation_{\pi_{\theta}}[R(\sigma)]$, where $R(\sigma)$ is the cumulative reward for a trajectory $\sigma = (s_0, a_0, r_0, s_1, a_1, r_1, ...)$ generated under the parameterized policy $\pi_{\theta}$. For gradient-based RL algorithms, the gradient of the objective function is derived as:
\[ \nabla_{\theta} J(\theta) =  \expectation_{\pi_{\theta}}[\nabla_{\theta}\log{\pi_{\theta}(a|s)}Q^{\pi}(s,a)]  \]
The parameter $\theta$ is updated using gradient ascent:
\[ \theta \leftarrow \theta + \alpha \nabla_{\theta} J(\theta) \]
where $\alpha \in (0,1]$ is the learning rate. Finally, we have the following theorem:
\begin{theorem}
    The optimal policy $\pi_{\theta}^*$ with parameters $\theta$ for the MDP $M$ will generate the paths that solve the optimization problem described in (\ref{eq:opt}).
\end{theorem}
In the next section, we will describe the construction of the environment for the MDP $M$. 

\section{Environment Design}
\label{sec:env_design}

To construct the environment for the MDP defined in the previous section, we used the well-known Gymnasium \citep{towers2024gymnasium} Python library. The environment takes the following inputs: the vertices of a 2-D polygon representing the field, the regions of interest (ROIs) that need to be visited, the initial positions of the robots, and the wind speed and velocity (defaults are zeros). The state and action space are defined using the \verb|gymnasium.spaces.Box| class (continuous and infinite). Let $p_{min}, p_{max}$ be the minimum and maximum position vectors in the 2-D space, and $v_{min}, v_{max}$ be the minimum and maximum velocities for each robot. Therefore, for $n$ robots and $m$ regions of interests, the state space $S$ is defined as,
\[  S = [p_{min}, p_{max}]^{n \times 2} \times [v_{min},v_{max}]^{n \times 2} \times \{0, 1,..., m\}    \]
Similarly, the action space is defined as,
\[  A = [f_{x_{min}}, f_{x_{max}}]^{n \times 1} \times [f_{y_{min}}, f_{y_{max}}]^{n \times 1}  \]

The transition function $\trans$ is defined using the dynamics above. We describe how to get the reward $R(s)$ based on the four reward functions $R_C(s), R_I(s), R_F(s)$ and $R_V(s)$, since it is the most important design aspect of the environment. If the new state $s'$ contains any collision between two robots, the reward $R(s)$ will be $-\infty$ and the current episode will be terminated. If $s'$ contains any location within the regions of interest $W$, then $R(s)$ is increased based on the number of locations present in $W$. If $s'$ has locations outside the environment $F$, then we decrease $R(s)$ significantly based on the number. If all locations in $s'$ are within the environment, we slightly decrease $r$ to keep the path lengths minimum. If $s'$ has locations that were visited previously, we moderately decrease $R(s)$ based on the number (if not, then we do not decrease). If all regions of interest have been visited, then the reward will be $+\infty$ and the current episode will be terminated. For full implementation details, readers are encouraged to check out the GitHub repository for this paper\footnote{\url{https://github.com/ch0t0n/MultiBotNav}}. Our environment is highly customizable, and the comments and explanations of each method can guide future researchers on how to modify them properly for their applications. We will generate variations of the environment and perform several experiments using different RL algorithms in the next section.

\section{Experiments}
\label{sec:experiments}


\begin{figure}[tbp]
    \centering
    \subfloat[Setting A (default)]{\includegraphics[width=0.48\linewidth]{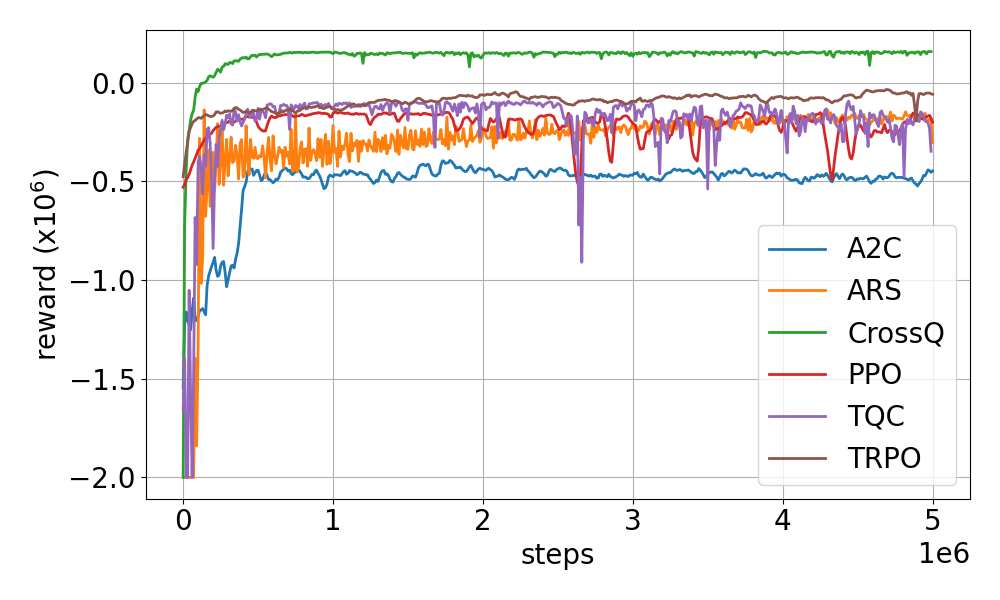}\label{fig:setting_a}}
    \hfill
    \subfloat[Setting B (random)]{\includegraphics[width=0.48\linewidth]{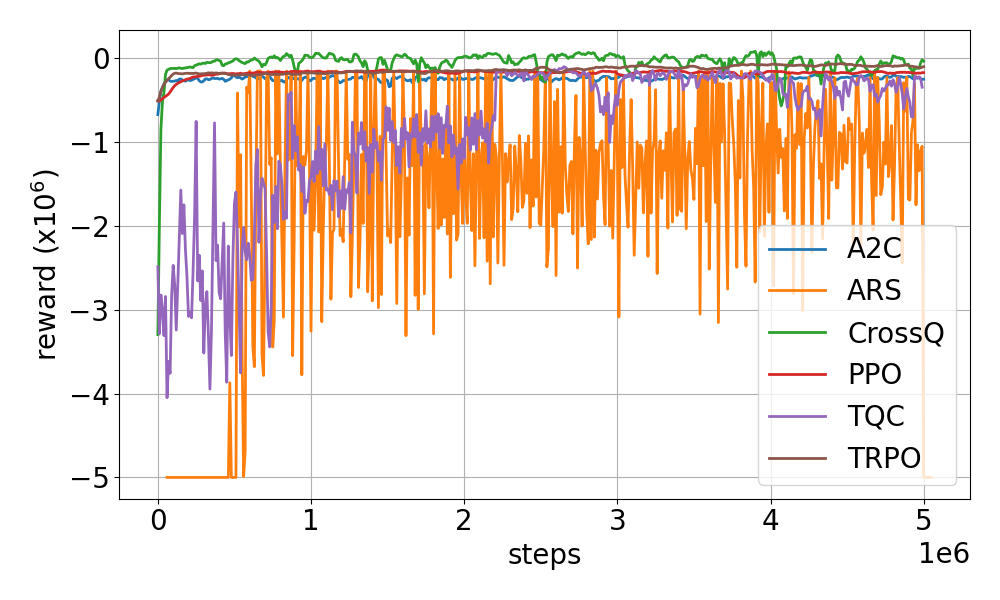}\label{fig:setting_b}} 
    \\ \vspace{-10pt}
    \subfloat[Setting C (transfer)]{\includegraphics[width=0.48\linewidth]{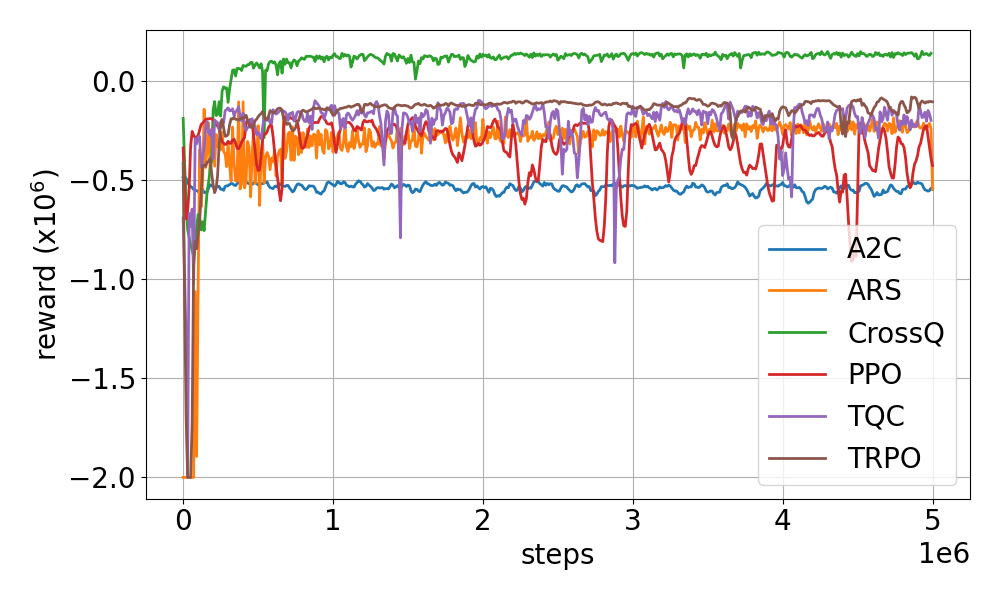}\label{fig:setting_c}}
    \hfill
    \subfloat[Optuna trials]{\includegraphics[width=0.48\linewidth]{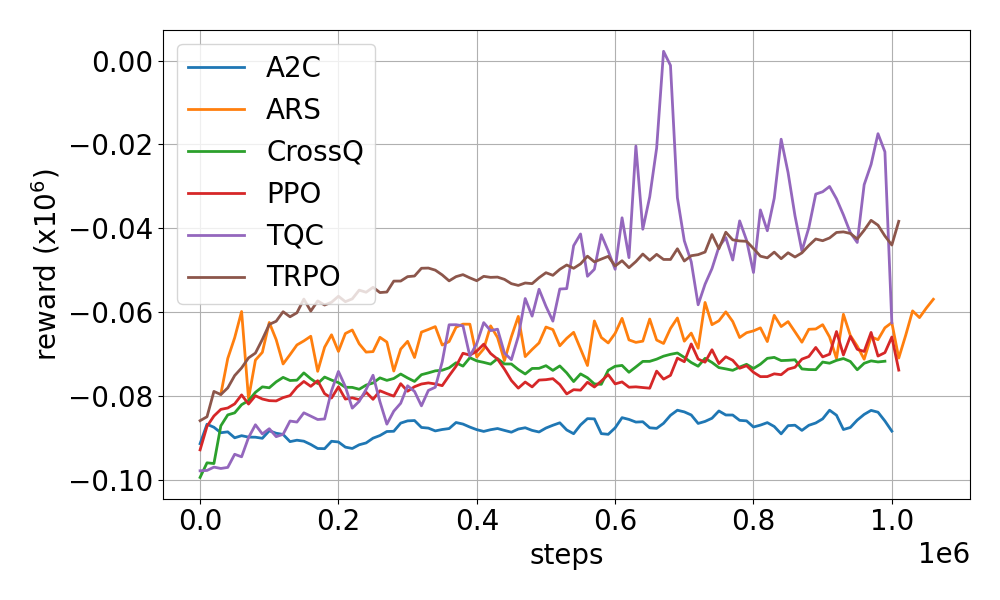}\label{fig:optuna_trials}}
    \caption{Plots (a), (b), and (c) show the mean episodic reward over all environment variations for settings A, B, and C, respectively. Plot (d) shows the mean episodic reward over 20 different Optuna trials in environment variation 1 during hyperparameter tuning.}
    \label{fig:algorithm_comparison}
\end{figure}



Experiments and simulations were conducted by crafting ten different variations (1-10) with different field shapes, robot positions, and regions of interest (ROIs). We used the Stable-Baselines3 and SB3-Contrib \cite{stable-baselines3} Python packages for measuring the performance of five state-of-the-art gradient-based deep RL algorithms (A2C \cite{mnih2016a2c}, TRPO \cite{schulman2015trust}, PPO \cite{schulman2017ppo}, TQC \cite{kuznetsov2020tqc}, CrossQ \cite{bhatt2024crossq}) and a finite-difference based algorithm (ARS \cite{mania2018ars}). Therefore, we have 6 algorithms $\times$ 10 environment variations = 60 experiments in total. We chose both on-policy (A2C, TRPO, PPO, ARS) and off-policy (CrossQ, TQC) RL methods for our experiments. The objective is to compare the learning efficiency of these RL algorithms for different environmental variations in a variety of training conditions. Their performances are measured based on cumulative rewards over each episode. The average reward threshold for the optimal policies of the environments is $0.1 \times 10^6$, which corresponds to optimal (or near-optimal) navigation paths. As \citet{henderson2018deep} mentioned, it is important to evaluate RL methods on a variety of hyperparameters. Therefore, we performed the experiments in three different settings:
\begin{itemize}
\item 
\textbf{Setting A}: Each algorithm is trained from scratch for all environment variations using the default hyperparameters given in Stable-Baselines3.

\item 
\textbf{Setting B}: Each algorithm is trained from scratch for all environment variations using hyperparameters selected by Optuna \cite{optuna_2019}.
 
\item 
\textbf{Setting C}: Each algorithm is trained on the first environment variation, and fine-tuned on the other nine variations through transfer learning.
\end{itemize}


We selected both common and model-specific hyperparameters for setting B. For a specific hyperparameter, we used the same range for all algorithms for consistency. The learning rate $\alpha \in [0.0001 - 0.05]$ is used by all algorithms. The discount factor $\gamma \in [0.90 - 0.99]$ is used in all gradient-based algorithms. The generalized advantage estimator $\lambda \in [0.90 - 1.0]$ is used for all algorithms other than ARS and TQC. The value function coefficient $\verb|vf_coef| \in [0.2 - 0.7]$, entropy coefficient $\verb|ent_coef| \in [0.0 - 0.05]$ and the maximum gradient clipping $\verb|max_grad_norm| \in [0.30 - 0.99]$ are only used in A2C, PPO and CrossQ. The $\verb|buffer_size| \in [1000-100000]$ is used on all off-policy algorithms (CrossQ, TQC). We performed 20 different trials in Optuna \cite{optuna_2019} for 1 million timesteps, each using environment variation 1 to tune the hyperparameters. For transfer learning, we first trained environment variation 1, and fine-tuned the trained model for environment variations 2-10. Unless otherwise specified, all experiments were run for 5 million timesteps.

\subsection{Results and Analysis}
Table~\ref{tab:alg_analysis} shows the results of the experiments for all three settings. For setting A, Fig.~\ref{fig:setting_a} shows the performance of each algorithm using the default hyperparameters, where it is evident that CrossQ outperformed all other algorithms by reaching the reward threshold within 500k timesteps. TRPO, PPO, and TQC have similar performances, but TRPO seems to be improving slowly. ARS reached a local optimum in nearly 4.9 million timesteps, but A2C was stuck in another local optimum lower than the others. The maximum reward for TRPO in setting A is higher than CrossQ, which confirms that CrossQ was also stuck on a local optimum very close to the global optimal reward, and the spike for the maximum reward for TRPO is not visible in the figure.

For setting B, Fig.~\ref{fig:optuna_trials} illustrates the results for performing 20 different trials on each algorithm in environment variation 1 using Optuna \cite{optuna_2019} to select the hyperparameters. Here, TQC and TRPO significantly outperformed other algorithms, indicating that we have found effective hyperparameters. Fig.~\ref{fig:setting_b} shows the performance of each algorithm using the selected hyperparameters. Here, CrossQ still performed slightly better than other algorithms. TRPO, PPO, and A2C were robust against changes in hyperparameters and performed similarly to setting A. TQC and ARS had very high variance at the beginning, but TQC was able to stabilize after 2 million timesteps. The high variance is because the optimal learning rate for environment variation 1 was too high for most of the other environments.

For setting C, Fig.~\ref{fig:setting_c} shows the performance of each algorithm during transfer learning. The performance is similar but slightly worse than setting A for all algorithms. We expected it to perform better than setting A, which could motivate us to use transfer learning to optimize the algorithm for specific variations instead of training it on the whole environment from scratch. In the future, we would like to explore how to perform transfer learning more efficiently, and fine-tune it for different environment variations using fewer timesteps than the full training.

\begin{table}[tbp]
\centering
\caption{Experiment results for each algorithm over all environments}
\resizebox{\linewidth}{!}{
\begin{tabular}{|c|c|c|c|c|c|c|c|c|c|c|c|c|c|}
\hline
\multirow{3}{*}{\textbf{Algorithm}} & \multicolumn{4}{c|}{\textbf{Setting A}} & \multicolumn{4}{c|}{\textbf{Setting B}} & \multicolumn{4}{c|}{\textbf{Setting C}} \\ \cline{2-13}
 & \textbf{Mean} & \textbf{SD} & \textbf{Max} & \textbf{Range} & \textbf{Mean} & \textbf{SD} & \textbf{Max} & \textbf{Range} & \textbf{Mean} & \textbf{SD} & \textbf{Max} & \textbf{Range} \\ 
 & $\times 10^{6}$ & $\times 10^{6}$ & $\times 10^{6}$ & $\times 10^{6}$  & $\times 10^{6}$ & $\times 10^{6}$ & $\times 10^{6}$ & $\times 10^{6}$ & $\times 10^{6}$ & $\times 10^{6}$ & $\times 10^{6}$ & $\times 10^{6}$ \\ \hline 
A2C     & -0.5160 & 0.2437 & -0.1002 & 6.6097 & -0.2436 & 0.0990 & -0.0820 & 2.3568 & -0.5381 & 0.1710 & -0.1817 & 1.2343 \\ \hline
PPO     & -0.2128 & 0.1846 & -0.0900 & 3.7044 & -0.1807 & 0.0669 & -0.0767 & 0.6550 & -0.3310 & 0.4485 & -0.0959 & 6.1574 \\ \hline
TRPO    & -0.0954 & 0.1157 & 0.1594  & 1.6792 & -0.1404 & 0.1011 & 0.0067  & 0.6613 & -0.1713 & 0.3700 & 0.1594  & 10.5443 \\ \hline
ARS     & -0.2706 & 1.4830 & -0.0802 & 29.7291 & -1.5345 & 4.3781 & -0.1408 & 19.7610 & -0.2916 & 1.5812 & -0.0703 & 29.6017 \\ \hline
CrossQ  & 0.1429  & 0.0990 & 0.1593  & 9.9617 & -0.0554 & 0.3986 & 0.1593  & 11.7615 & 0.1190  & 0.1113 & 0.1595  & 1.6780 \\ \hline
TQC     & -0.1977 & 0.5731 & 0.1326  & 21.1364 & -0.8592 & 2.7137 & 0.0142  & 18.5191 & -0.2116 & 0.3011 & -0.0215 & 6.8070 \\ \hline

\end{tabular}}
\label{tab:alg_analysis}
\end{table}

\subsection{Simulation}
\begin{figure}[tbp]
    \centering
    \subfloat[Initial]{\includegraphics[width=0.2\textwidth, height=0.12\textwidth]{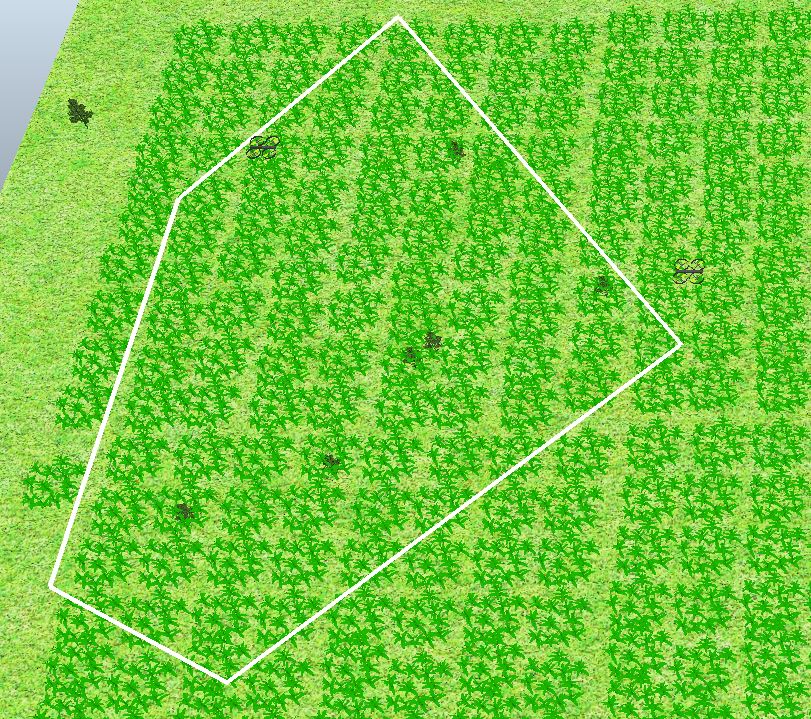}\label{fig:coppelia_set1_i}}
    \hfil
    \subfloat[Final]{\includegraphics[width=0.2\textwidth, height=0.12\textwidth]{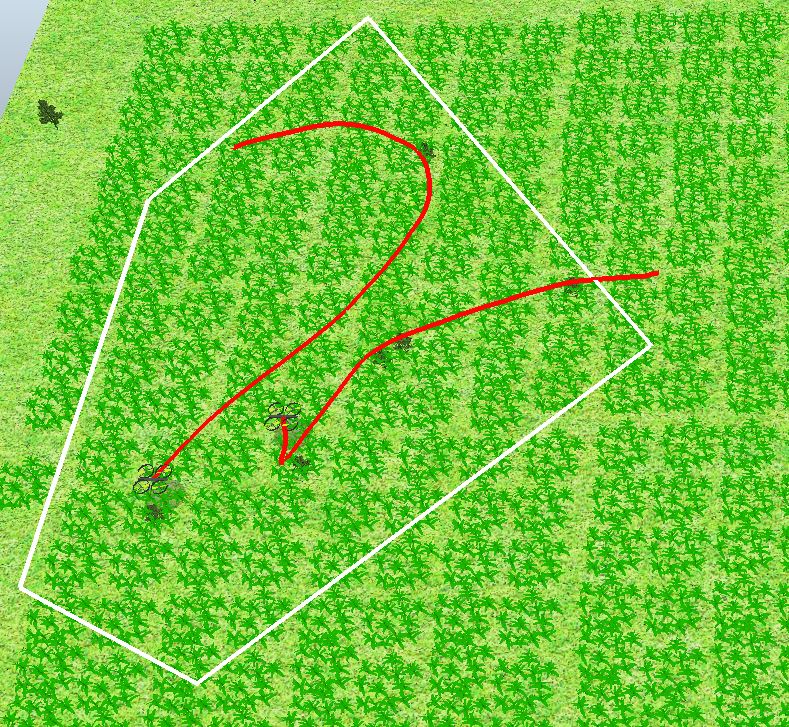}\label{fig:coppelia_set1_f}} 
    \hfil
    \subfloat[Initial]{\includegraphics[width=0.2\textwidth, height=0.12\textwidth]{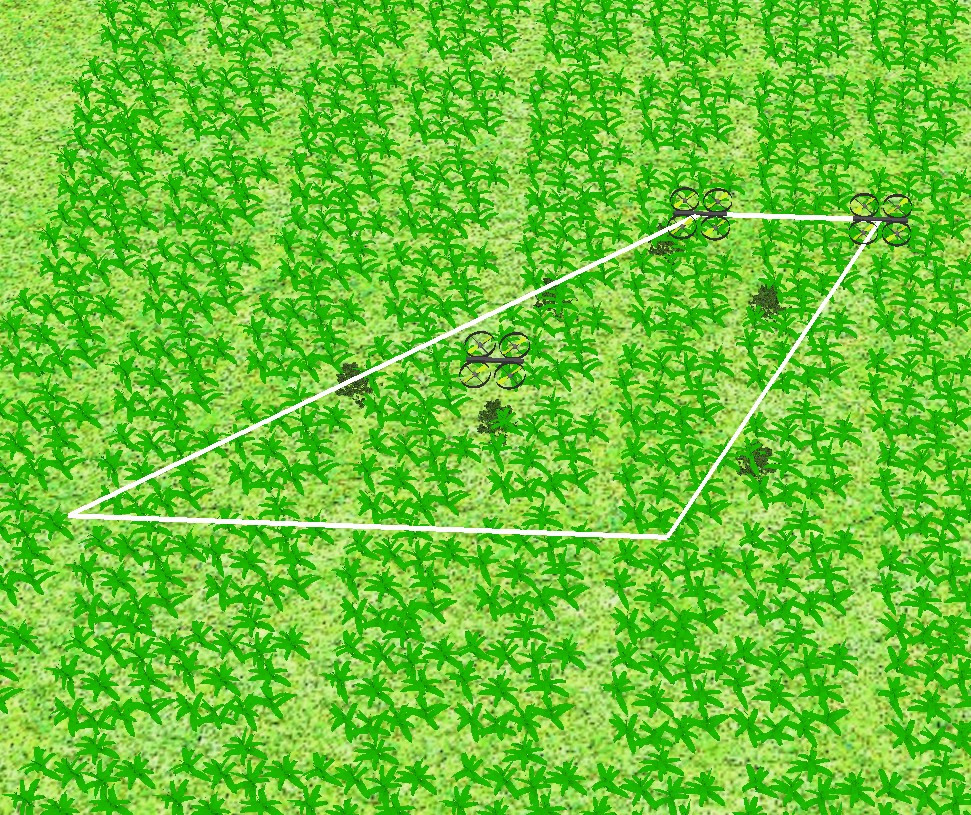}\label{fig:coppelia_set3_i}}
    \hfil
    \subfloat[Final]{\includegraphics[width=0.2\textwidth, height=0.12\textwidth]{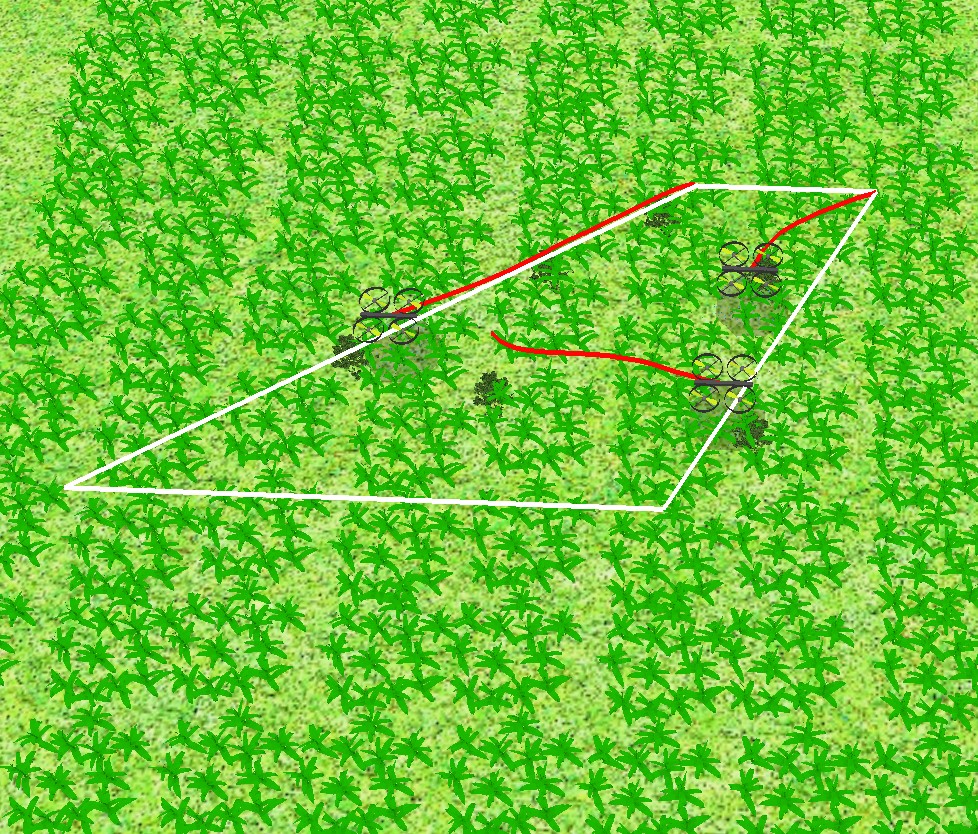}\label{fig:coppelia_set3_f}}    
    \caption{Simulation of environment variation 1 (a, b) and 3 (c, d) in CoppeliaSim using two and three uncrewed aerial vehicles (UAVs) respectively.}
    \label{fig:coppelia_sim} \vspace{-20pt}
\end{figure}

We used CoppeliaSim \cite{coppeliaSim2013} (previously known as V-Rep \cite{vrep2013}) to develop a 3D scene of a cornfield with weeds in certain locations. The task is to spray herbicides on the weed locations using multiple uncrewed aerial vehicles (UAVs). CoppeliaSim is a widely used simulator due to its real-time simulation capabilities and its ability to create dynamic environments that closely model real-world scenarios. The physics of the surrounding environment was computed by the PyBullet \cite{coumans2021pybullet} engine. We use the ZeroMQ \cite{ZeroMQ} remote API client to connect our custom Gymnasium environment with CoppeliaSim. Since our environment was developed using the dynamics for navigation in 2-D space, we operated the drones at a fixed height (0.35 m) above the ground in the 3D agricultural scene. Fig. \ref{fig:coppelia_sim} shows the simulation of environment variations 1 and 3 with two and three UAVs, respectively, by running inference on the models trained using the CrossQ \cite{bhatt2024crossq} algorithm. The starting positions of the UAVs and weed locations (\textcolor{darkgreen}{dark green} plants) are shown in Fig.~\ref{fig:coppelia_set1_i} and Fig.~\ref{fig:coppelia_set3_i}. The final states are shown in Fig.~\ref{fig:coppelia_set1_f} and Fig.~\ref{fig:coppelia_set3_f}, where the \textcolor{red}{red} lines are the spraying paths covering all weed locations. The white lines are the field boundary. The dynamics of the drones are computed by the simulator based on the sequence of actions taken by the trained model.

\begin{figure}[tbp]
    \centering
    \subfloat[$v_a=0.1$]{\includegraphics[width=0.15\textwidth, height=0.12\textwidth]{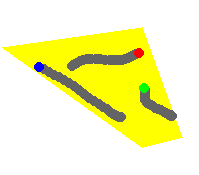}\label{fig:env3p_0.1}} 
    \hfil
    \subfloat[$v_a=0.2$]{\includegraphics[width=0.15\textwidth, height=0.12\textwidth]{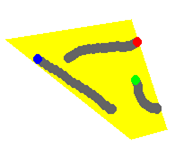}\label{fig:env3p_0.2}}
    \hfil
    \subfloat[$v_a=0.3$]{\includegraphics[width=0.15\textwidth, height=0.12\textwidth]{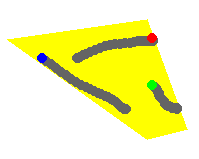}\label{fig:env3p_0.3}}
    \hfil
    \subfloat[$v_a=0.4$]{\includegraphics[width=0.15\textwidth, height=0.12\textwidth]{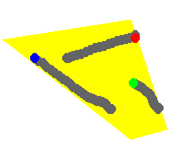}\label{fig:env3p_0.4}}
    \hfil
    \subfloat[$v_a=0.5$]{\includegraphics[width=0.15\textwidth, height=0.12\textwidth]{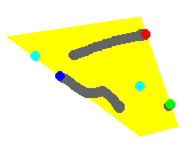}\label{fig:env3p_0.5}}
    \\
    \subfloat[$v_a=0.1$]{\includegraphics[width=0.15\textwidth, height=0.12\textwidth]{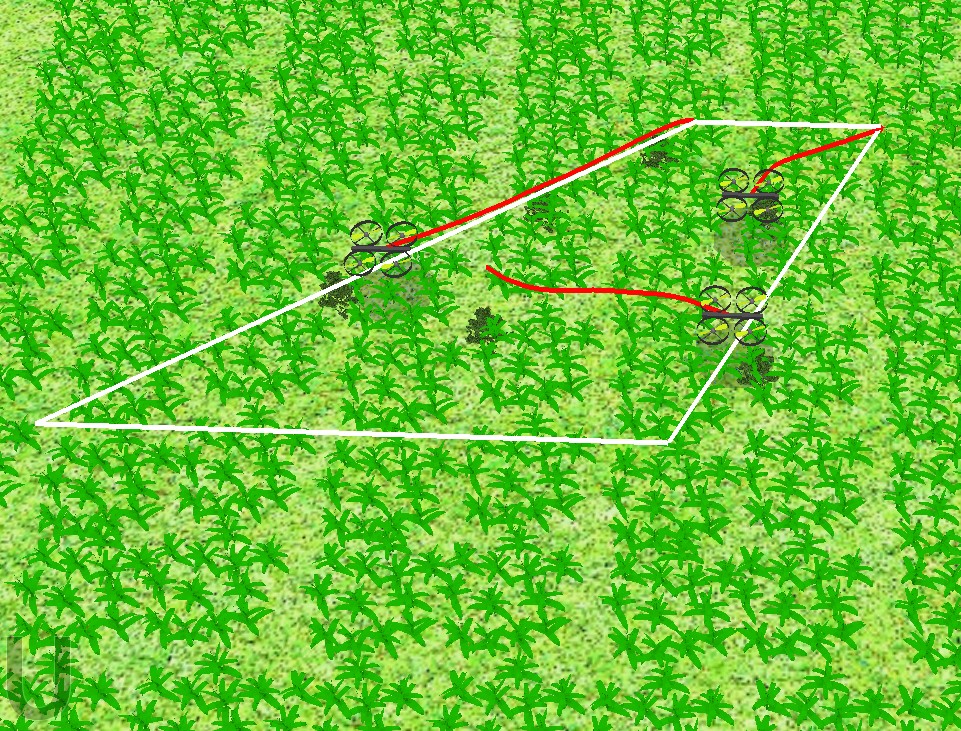}\label{fig:env3s_0.1}} 
    \hfil
    \subfloat[$v_a=0.2$]{\includegraphics[width=0.15\textwidth, height=0.12\textwidth]{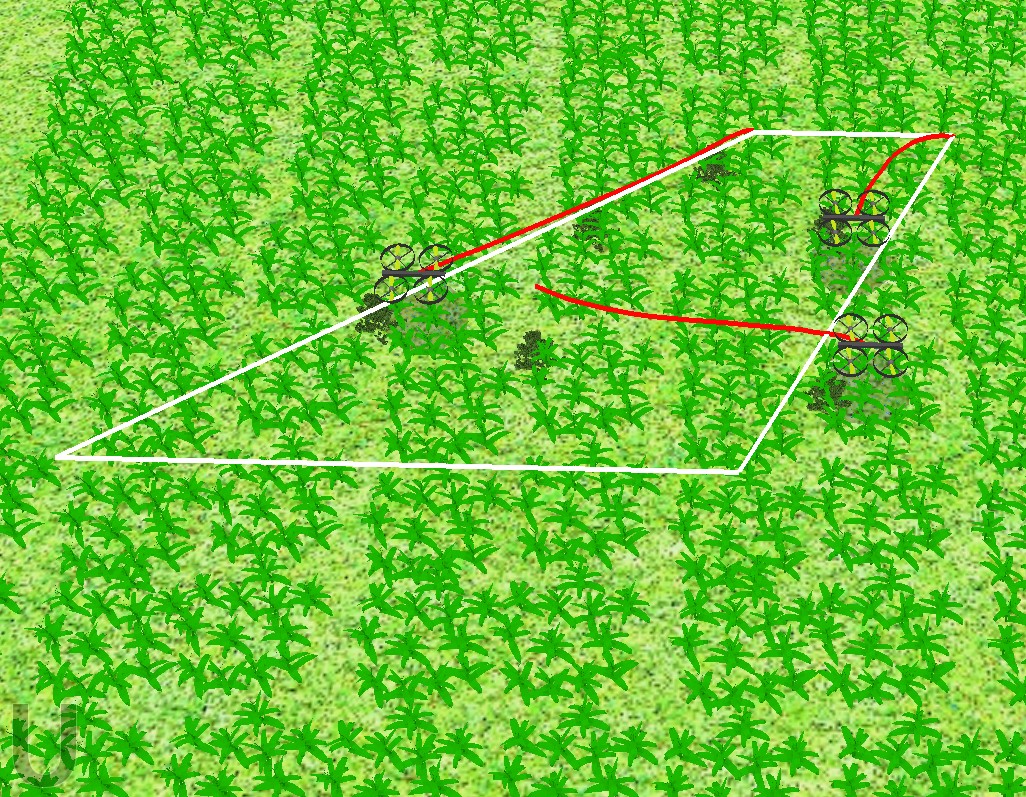}\label{fig:env3s_0.2}}
    \hfil
    \subfloat[$v_a=0.3$]{\includegraphics[width=0.15\textwidth, height=0.12\textwidth]{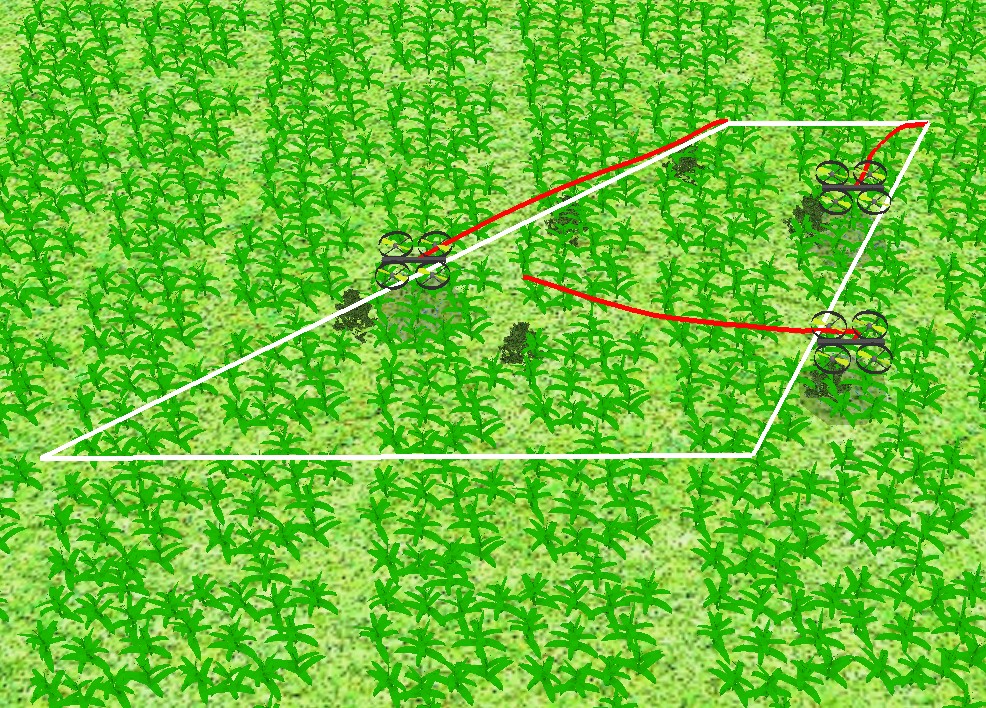}\label{fig:env3s_0.3}}
    \hfil
    \subfloat[$v_a=0.4$]{\includegraphics[width=0.15\textwidth, height=0.12\textwidth]{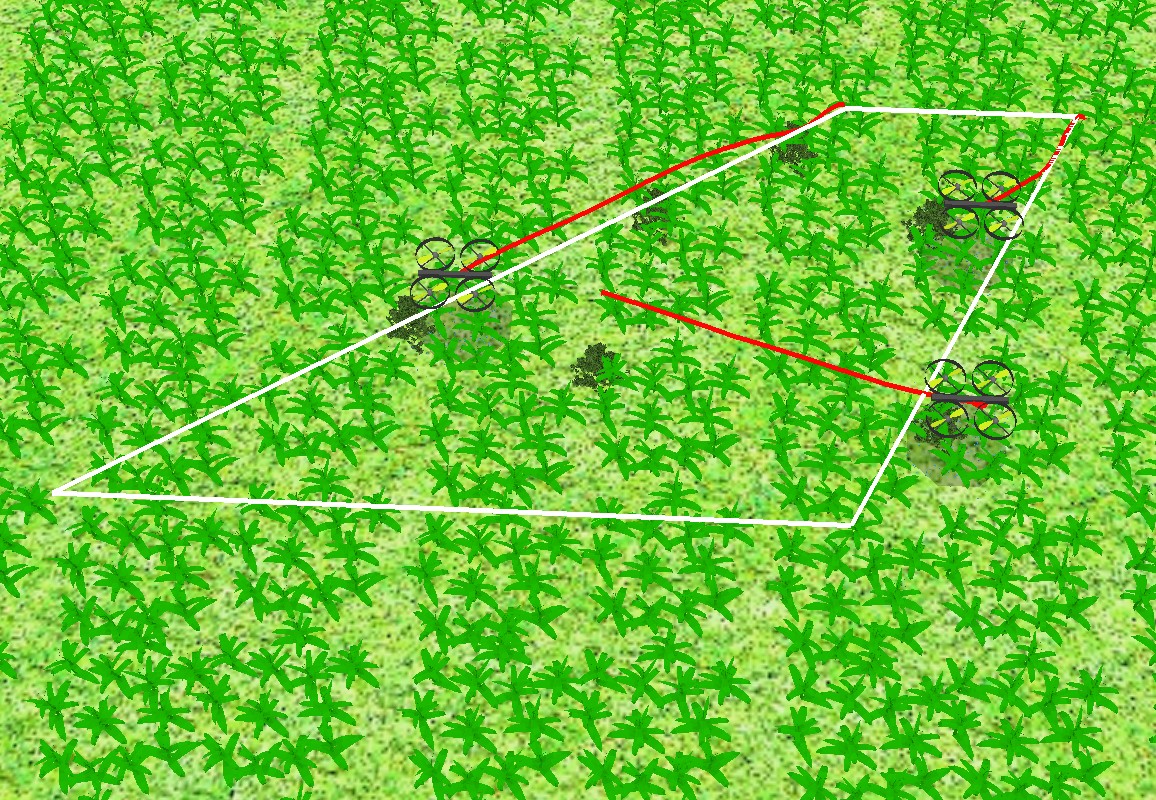}\label{fig:env3s_0.4}}
    \hfil
    \subfloat[$v_a=0.5$]{\includegraphics[width=0.15\textwidth, height=0.12\textwidth]{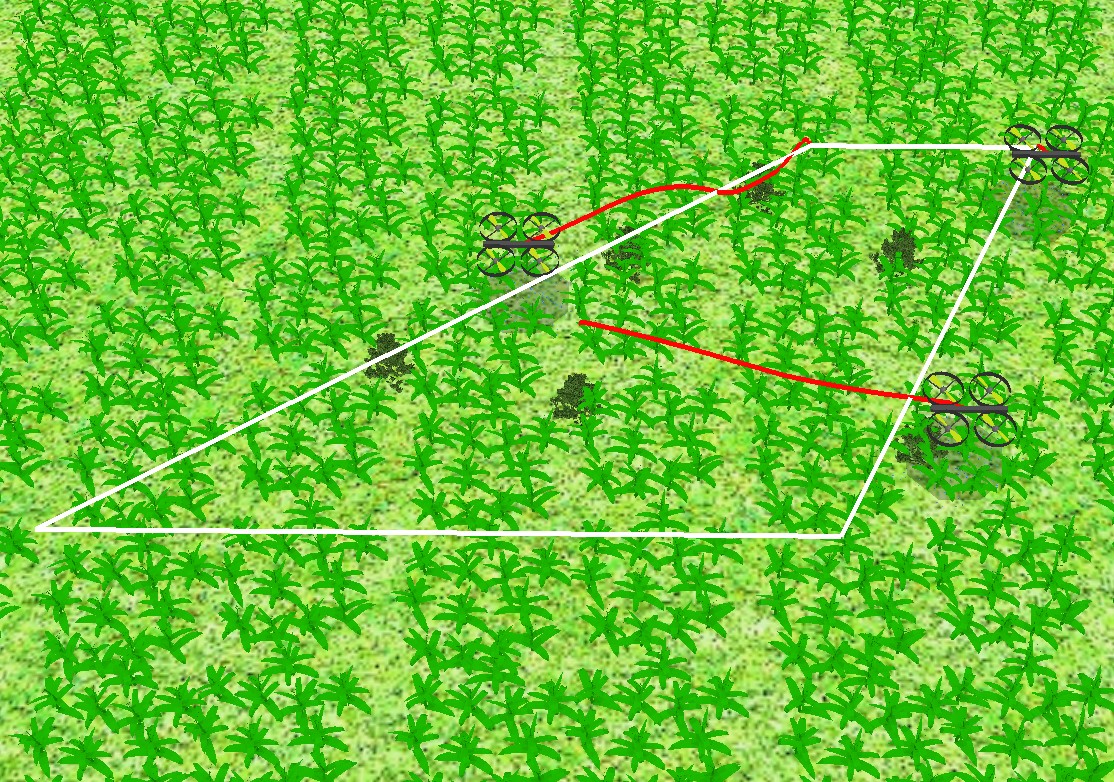}\label{fig:env3s_0.5}}   
    \caption{Simulation of the windy environment with angle $\beta_a=30$° and variable speed $v_a$ using Pygame (a-d) and CoppeliaSim (e-j). The trained model was able to find the optimal path for wind speed up to $0.4$ unit per second.}
    \label{fig:windy_sim} \vspace{-8pt}
\end{figure}

To measure the robustness, we ran inference of the learned model for environment 3 in windy conditions. The model was trained using the CrossQ algorithm with no winds, and the inference was run with the wind blowing with speed $v_a$ and angle $\beta_a$. Fig. \ref{fig:windy_sim} shows the results where the robots were able to find the optimal spraying paths where $v_a < 0.5$. The robustness depends on the maximum and minimum speeds of the UAVs by taking any action. We use $v_{clip} = \pm5$ unit per second for the maximum and minimum speed that prevents a UAV from gaining unreasonable speed that cannot be visualized in the simulators. Therefore, the engineers can use the trained model to generate the optimal paths in uncertain and windy environments if $ -\frac{v_{clip}}{10} < v_a < \frac{v_{clip}}{10}$ holds.

\vspace{-5pt}
\section{Conclusion}
\label{sec:conclusion}
We explored deep RL methods for multi-robot navigation in continuous state and action spaces with environmental uncertainties. Our environment is highly customizable, which can be used with any polygonal field with up to 10 regions of interest. The experimental results and simulations show that despite having large and continuous state and action spaces, we can find optimal navigation paths within reasonable training periods for any variation. In the future, we would like to explore 3-D path planning methods for UAVs by varying their heights, and adding more constraints such as failure or disconnection of some robots while finding the optimal path for the remaining active robots.
\section{Limitations}
\label{sec:limitations}

This work is an improvement of a previous work where the state and action spaces were discrete. We plan to continue our work on this project for different noisy environments with different ground and aerial robots. The multi-robot navigation was performed in a continuous space with 2-D locations, which can be extended to 3-D locations for the navigation of aerial robots. The maximum path or episode length (correlated to flight time) for any robot is fixed and assumed to be 1000, which was enough for every environment variation to visit all regions of interest (each episode terminates after 1000 steps). But if the episode length (or flight time) is not enough to visit all locations, then we may get a sub-optimal path, for which further experiments need to be performed. All robots are identical and follow the same dynamics. The dynamics given in section \ref{sec:problem} are relatively simple, and the robots are assumed to be omnidirectional that can move in any direction. The environment can be easily modified to have other 2-D dynamics such as Dubin's path \cite{dubins1957}, and we implemented it for coverage path planning (CPP) in one of our previous works \cite{choton2023icra}. 

The sample complexity of the state space increases exponentially as the number of robots increases. We tested our environment with up to 7 robots. As Stable-Baselines3 requires the state and action space to be exactly the same for transfer learning, we picked the same number of ROIs (6) and the same number of robots (3) for the transfer learning experiments. However, the locations of the ROIs were randomized. Even though the learned model can reach a reward threshold within a reasonable time, the absolute maximum reward cannot be guaranteed. Hence, the optimal path can change slightly. For example, environment variation 1 has the reward of 109995 for the optimal path, but the learned model got the reward of 109980 (15 less than optimal). The hyperparameters for setting B were selected based on the performance in environment variation 1, and hence not completely random. Only 20 trials were performed with Optuna to select the hyperparameters, so future work could look into increasing performance and robustness of the results by tuning for more trials. All experiments were also run on one seed (33), so robustness of results could also be improved by running additional experiments across multiple seeds. Some critical safety parameters such as collision avoidance were implemented in the design of the reward function, but that doesn't guarantee the robots to avoid collisions altogether for the learned model. We may need to perform additional safety verification tests for the learned models.

For the simulation, the agricultural scene was made with built-in quadcopter models from CoppeliaSim. The dynamics of each quadcopter (controlling the four motors) are automatically computed by the simulator for a given flight path. However, it is also possible to change the dynamics by editing the scripts of each quadcopter. The remote API used for connecting the simulator with our Python scripts can have a very small communication delay. The robots are controlled centrally by a host, and there is no communication among themselves. We performed additional simulations using autonomous helicopters and balancing robots, but didn't include those results due to page limits. They are available in the supplementary materials.

Despite these limitations, we believe that our work will guide researchers on how to improve the implementation of RL in real-time systems in each development stage, from formally defining the MDP to controlling physical robots in uncertain and noisy environments.

\clearpage

\bibliography{biblio}  

\end{document}




\section{Additional Results and Simulations}
Fig.~\ref{fig:all_envs} illustrates the ten environment variations with three robots and six regions of interest (ROIs) that were used for the experiments. The vertices of the boundaries were selected by taking random 2-D points between (0,0) and (1000,1000). The initial robot positions and ROIs were selected by taking random 2-D points inside the boundaries. Table~\ref{tab:random_hyperparameters} shows the values of the hyperparameters selected by Optuna by running 20 trials on environment variation 1. Fig.~\ref{fig:hyperparameters} compares the performance on settings A and B for each algorithm. Here, only A2C performed better for setting B, and all other algorithms performed better on setting A. 

Fig.~\ref{fig:transfer} compares the  performance on settings A and C for each algorithm. The transfer learning approach performed very similar to non-transfer learning. Table~\ref{tab:setting_a_envs}, \ref{tab:setting_b_envs}, and \ref{tab:setting_c_envs} shows the performance of each environment variation over all algorithms for settings A, B and C respectively. Here, variation 3 outperformed all other variations because it has the least area, so finding the optimal path required fewer iterations.

Additional simulations were also performed using autonomous helicopters and balancing robots. Fig.~\ref{fig:heli_sim} shows the simulation of environment variations 1 and 3 using autonomous helicopters. Fig.~\ref{fig:ball_sim} shows the simulation of environment variations 1 and 3 using autonomous balancing robots. The dynamics of these robots were automatically computed by the simulator for a given path. For the balancing robots, we had to use a smaller clipping value for maximum and minimum speeds ($v_{clip}=2$) to prevent the robots from falling.

\begin{figure}[hbp]
    \centering
    \includegraphics[width=1\linewidth]{figures/all_environments.png}
    \caption{All environment variations. The initial robot positions are shown in \textcolor{red}{red} and the ROIs are shown in \textcolor{darkgreen}{green}. }
    \label{fig:all_envs}
\end{figure}

\begin{figure}[hbp]
    \centering
    \subfloat[Initial]{\includegraphics[width=0.24\textwidth, height=0.18\textwidth]{figures/heli_env1_init.jpg}\label{fig:heli_set1_i}}
    \hfill
    \subfloat[Final]{\includegraphics[width=0.24\textwidth, height=0.18\textwidth]{figures/heli_final_env1.jpg}\label{fig:heli_set1_f}} 
    \hfill
    \subfloat[Initial]{\includegraphics[width=0.24\textwidth, height=0.18\textwidth]{figures/heli_env3_init.jpg}\label{fig:heli_set3_i}}
    \hfill
    \subfloat[Final]{\includegraphics[width=0.24\textwidth, height=0.18\textwidth]{figures/heli_final_env3.jpg}\label{fig:heli_set3_f}}    
    \caption{Simulation of environment variation 1 (a, b) and 3 (c, d) in CoppeliaSim using two and three autonomous helicopters respectively.}
    \label{fig:heli_sim} \vspace{-8pt}
\end{figure}

\begin{figure}[tbp]
    \centering
    \subfloat[Initial]{\includegraphics[width=0.24\textwidth, height=0.18\textwidth]{figures/ball_env1_init.jpg}\label{fig:ball_set1_i}}
    \hfill
    \subfloat[Final]{\includegraphics[width=0.24\textwidth, height=0.18\textwidth]{figures/ball_env1_final.jpg}\label{fig:ball_set1_f}} 
    \hfill
    \subfloat[Initial]{\includegraphics[width=0.24\textwidth, height=0.18\textwidth]{figures/ball_env3_init.jpg}\label{fig:ball_set3_i}}
    \hfill
    \subfloat[Final]{\includegraphics[width=0.24\textwidth, height=0.18\textwidth]{figures/ball_env3_final.jpg}\label{fig:ball_set3_f}}    
    \caption{Simulation of environment variations 1 (a, b) and 3 (c, d) in CoppeliaSim using two and three autonomous balancing robots respectively.}
    \label{fig:ball_sim} \vspace{-8pt}
\end{figure}

\begin{figure}[tbp]
    \centering
    \subfloat[A2C]{\includegraphics[width=0.33\linewidth]{figures/a2c_hyper2.png}\label{fig:a2c_h}}
    \hfill
    \subfloat[PPO]{\includegraphics[width=0.33\linewidth]{figures/ppo_hyper2.png}\label{fig:ppo_h}} 
    \hfill
    \subfloat[TRPO]{\includegraphics[width=0.33\linewidth]{figures/trpo_hyper2.png}\label{fig:trpo_h}}
    \hfill
    \subfloat[ARS]{\includegraphics[width=0.33\linewidth]{figures/ars_hyper2.png}\label{fig:ars_h}}
    \hfill
    \subfloat[CrossQ]{\includegraphics[width=0.33\linewidth]{figures/crossq_hyper3.png}\label{fig:crossq_h}} 
    \hfill
    \subfloat[TQC]{\includegraphics[width=0.33\linewidth]{figures/tqc_hyper2.png}\label{fig:setting_b}}
    \caption{Comparison of performance on settings A and B (default and selected hyperparameters).}
    \label{fig:hyperparameters}
\end{figure}

\begin{figure}[tbp]
    \centering
    \subfloat[A2C]{\includegraphics[width=0.33\linewidth]{figures/a2c_trans.png}\label{fig:a2c_trans}}
    \hfill
    \subfloat[PPO]{\includegraphics[width=0.33\linewidth]{figures/ppo_trans.png}\label{fig:ppo_trans}} 
    \hfill
    \subfloat[TRPO]{\includegraphics[width=0.33\linewidth]{figures/trpo_trans.png}\label{fig:trpo_trans}}
    \hfill
    \subfloat[ARS]{\includegraphics[width=0.33\linewidth]{figures/ars_trans.png}\label{fig:ars_trans}}
    \hfill
    \subfloat[CrossQ]{\includegraphics[width=0.33\linewidth]{figures/crossq_trans.png}\label{fig:crossq_trans}} 
    \hfill
    \subfloat[TQC]{\includegraphics[width=0.33\linewidth]{figures/tqc_trans.png}\label{fig:tqc_trans}}
    \caption{Comparison of performance on settings A and C (non-transfer and transfer learning).}
    \label{fig:transfer}
\end{figure}

\begin{table}[tbp]
\centering
\caption{Hyperparameter values selected by Optuna trials.}
\begin{tabular}{|c|c|c|c|c|c|c|}
\hline
\textbf{Hyperparameter} & \textbf{A2C} & \textbf{TRPO} & \textbf{PPO} & \textbf{ARS} & \textbf{CrossQ} & \textbf{TQC} \\ \hline
$\gamma$                   & 0.974        & 0.989         & 0.968        & -            & 0.953           & 0.901        \\ \hline
$\lambda$         & 0.938        & 0.931         & 0.959        & -            & 0.959           & -            \\ \hline
$\alpha$          & 0.0007       & 0.0001        & 0.0001       & 0.019        & 0.0013          & 0.0001       \\ \hline
\verb|vf_coef|                & 0.632        & -             & 0.696        & -            & 0.531           & -            \\ \hline
\verb|ent_coef|               & 0.0001       & -             & 0.015        & -            & 0.0248          & -            \\ \hline
\verb|max_grad_norm|         & 0.674        & -             & 0.983        & -            & 0.6072          & -            \\ \hline
\verb|buffer_size|            & -            & -             & -            & -            & 95000           & 100000       \\ \hline
\end{tabular}
\label{tab:random_hyperparameters}
\end{table}

\begin{table}[hbp]
\centering
\caption{Experiment results for setting A averaged over all algorithms for each environment.}
\begin{tabular}{|c|c|c|c|c|c|}
\hline
\textbf{Environment} & \textbf{Mean} & \textbf{SD} & \textbf{Max} & \textbf{Range} \\ \hline
1   & -223295  & 1098396  & 158762.8 & 29615673 \\ \hline
2   & -294879  & 483046.3 & 156239.2 & 27665757 \\ \hline
3   & -33008.1 & 226538.5 & 159370   & 20694206 \\ \hline
4   & -111742  & 534880.6 & 159100   & 29968474 \\ \hline
5   & -238711  & 1231828  & 158368.9 & 29692261 \\ \hline
6   & -209736  & 1411111  & 158555.5 & 29501284 \\ \hline
7   & -79876.9 & 403011.2 & 159108.7 & 24397489 \\ \hline
8   & -333719  & 1910784  & 158948.8 & 29649277 \\ \hline
9   & -225092  & 406512.2 & 156358   & 27069268 \\ \hline
10  & -438714  & 754514.7 & 156351   & 26740725 \\ \hline
\end{tabular}
\label{tab:setting_a_envs}
\end{table}

\begin{table}[htbp]
\centering
\caption{Experiment results for setting B averaged over all algorithms for each environment.}
\begin{tabular}{|c|c|c|c|c|c|}
\hline
\textbf{Environment} & \textbf{Mean} & \textbf{SD} & \textbf{Max} & \textbf{Range} \\ \hline
1  & -297279  & 1237320  & 158800   & 19668936 \\ \hline
2  & -462186  & 1867179  & 158768.2 & 19714922 \\ \hline
3  & -358918  & 944604.9 & 159070   & 20060880 \\ \hline
4  & -514652  & 1480433  & 159100   & 20055770 \\ \hline
5  & -383453  & 1673897  & 159070   & 19816174 \\ \hline
6  & -421029  & 1922726  & 158506.6 & 19756919 \\ \hline
7  & -458493  & 1642766  & 159250   & 20015582 \\ \hline
8  & -341326  & 1648950  & 158677.9 & 19868722 \\ \hline
9  & -436148  & 1624478  & -27246.9 & 19050871 \\ \hline
10 & -1977281 & 5308743  & -23886.3 & 19056754 \\ \hline
\end{tabular}
\label{tab:setting_b_envs}
\end{table}

\begin{table}[htbp]
\centering
\caption{Experiment results for setting C averaged over all algorithms for each environment.}
\begin{tabular}{|c|c|c|c|c|c|}
\hline
\textbf{Environment} & \textbf{Mean} & \textbf{SD} & \textbf{Max} & \textbf{Range} \\ \hline
1   & -610337  & 1985139  & 158473   & 19668609 \\ \hline
2   & -323476  & 483077.6 & 364.4    & 23883158 \\ \hline
3   & 32867.09 & 264073   & 159490   & 22615808 \\ \hline
4   & -187699  & 484186.6 & 158882.5 & 28121787 \\ \hline
5   & -322870  & 1566003  & 158569.3 & 29830501 \\ \hline
6   & -464557  & 1344092  & 158468.2 & 29484164 \\ \hline
7   & -138294  & 349680.2 & 158952.1 & 28090986 \\ \hline
8   & -449114  & 2209255  & 158823.1 & 29628301 \\ \hline
9   & -252253  & 372374.1 & 157041.1 & 27165805 \\ \hline
10  & -624790  & 326000.2 & -8731    & 1716212  \\ \hline
\end{tabular}
\label{tab:setting_c_envs}
\end{table}